\documentclass[pdflatex,sn-nature]{sn-jnl}

\usepackage{graphicx}%
\usepackage{multirow}%
\usepackage{amsmath,amssymb,amsfonts}%
\usepackage{amsthm}%
\usepackage{mathrsfs}%
\usepackage[title]{appendix}%
\usepackage{xcolor}%
\usepackage{textcomp}%
\usepackage{manyfoot}%
\usepackage{booktabs}%
\usepackage{algorithm}%
\usepackage{algorithmicx}%
\usepackage{algpseudocode}%
\usepackage{listings}%
\theoremstyle{thmstyleone}%

\theoremstyle{thmstyletwo}%

\theoremstyle{thmstylethree}%

\definecolor{chcolor}{RGB}{46, 179, 81}

\begin{document}

\title{Investigating Robot Control Policy Learning for Autonomous X-ray-guided Spine Procedures}

\author*[1,2]{\fnm{Florence} \sur{Klitzner}}\email{florence.klitzner@tum.de}
\author[1]{\fnm{Blanca Inigo} \sur{Romillo}}
\author[1]{\fnm{Benjamin D.} \sur{Killeen}}
\author[1]{\fnm{Lalithkumar} \sur{Seenivasan}}
\author[1]{\fnm{Michelle} \sur{Song}}
\author[3]{\fnm{Rebecca} \sur{Choi}}
\author[3]{\fnm{Majid} \sur{Khan}}
\author[1]{\fnm{Axel} \sur{Krieger}}
\author[1]{\fnm{Mathias} \sur{Unberath}}

\affil[1]{\orgname{Johns Hopkins University}, \orgaddress{\city{Baltimore}, \postcode{21218}, \state{MD}, \country{USA}}}
\affil[2]{\orgname{Technical University of Munich}, \orgaddress{\city{Munich}, \postcode{80333}, \state{BY}, \country{Germany}}}
\affil[3]{\orgname{Johns Hopkins School of Medicine}, \orgaddress{\city{Baltimore}, \postcode{21205}, \state{MD}, \country{USA}}}

\abstract{

\textbf{Purpose:} Imitation learning-based robot control policies are enjoying renewed interest in video-based robotics. However, it remains unclear whether this approach applies to X-ray-guided procedures, such as spine instrumentation, with sparse inputs. We examine the feasibility, opportunities and challenges for imitation policy learning in bi-plane-guided cannula insertion. 

\textbf{Method:} We develop an \emph{in silico} sandbox for scalable, automated simulation of X-ray-guided spine procedures with a high degree of realism. We curate a dataset of correct trajectories and corresponding bi-planar X-ray sequences that emulate the stepwise alignment of providers. We then train imitation learning policies for planning and open-loop control that iteratively align a cannula in a vertebroplasty setting solely based on visual information. This precisely controlled setup offers insights into limitations and capabilities of this method. 

\textbf{Results:} Our policy succeeded on the first attempt in 68.5\% of cases, maintaining safe intra-pedicular trajectories across diverse vertebral levels. The policy transferred to complex anatomy, including fractures, as well as varied anatomies and initializations. Rollouts on real X-ray indicate that partial sim-to-real transfer with plausible trajectories is possible.

\textbf{Conclusion:} While these preliminary results are promising, we also identify limitations, especially in entry point precision. The current results present a clear benchmark for future efforts, while with more robust priors and domain knowledge, such models may provide a foundation for future efforts toward lightweight and CT-free robotic intra-operative spinal navigation.
}

\keywords{Image-guided surgery, Imitation Learning, Medical Robotics, Spinal Procedures}

\maketitle

\section{Introduction}

In fluoroscopy-guided spinal procedures such as vertebroplasty surgeons often rely solely on bi-plane X-ray imaging to plan and execute safe instrument trajectories. Interpreting these 2D projections to navigate complex 3D anatomy remains a demanding task, requiring substantial clinical experience~\cite{chen_risk_2020, bermejo_identification_2014, rampersaud_radiation_2000}. To assist with this, CT-based trajectory planning has been extensively studied, with methods like statistical shape models (SSMs) and deep learning achieving expert-level performance in anatomical targeting~\cite{vijayan_automatic_2019, siemionow_autonomous_2021}. Executing on these plans currently requires extensive specialized infrastructure, hardware and workflow adjustments~\cite{siemionow_autonomous_2021, huang_current_2021}. In this context the recent rapid progress in imitation learning (IL) for robot control has seen a renewed interest, due to its potential in enabling smaller low-cost systems with high environmental flexibility and performance. Transformer-based policies such as Action Chunking with Transformers (ACT)~\cite{zhao_learning_2023} have introduced architectures capable of predicting smooth, temporally coherent action sequences, enabling effective long-horizon execution of complex manipulation tasks. Domain-specific extensions, such as Surgical Robot Transformers (SRT)~\cite{kim_surgical_2024, kim_srt-h_2025}, have adapted this formulation to the surgical domain. These models support rollout on multi-arm systems, such as the Da Vinci robotic surgical system, demonstrating robust performance for fine-grained workflows including tissue manipulation and needle handover~\cite{kim_surgical_2024}. Their ability to learn structured policies from visual feedback highlights the generality and flexibility of transformer-based IL for both autonomous and assistive surgical robotics.

Crucially these advances in IL are powered by 3D scene understanding from multiple 2D camera views to achieve their task. Instead of relying purely on 3D data or registration, this could enable adapting IL policies toward 2D medical imaging problems, like fluoroscopy guidance. Though relying on limited 2D X-ray views offers only sparse and ambiguous image cues, making the task substantially more difficult. In particular, transpedicular cannula insertion requires precise 3D reasoning and scene understanding based solely on 2D projections. Highly realistic simulation frameworks, such as DeepDRR~\cite{unberath_deepdrr_2018}, enable large-scale, controlled training and testing of models in the X-ray domain, and have shown real sim-to-real transfer~\cite{unberath_deepdrr_2018, gao_synthetic_2023, killeen2023silico, killeen2023pelphix, killen2025fluorosam}. This makes it feasible to train and evaluate planning policies in this new modality without requiring manually annotated clinical datasets or real-world data collection. 

Our work provides a controlled investigation into the viability and limitations of creating vision-based trajectory planning systems for X-ray guided spinal procedures that mimic a surgeons' intra-operative stepwise decision-making. We rely on realistic simulation \cite{unberath_deepdrr_2018} to enable experiments across a wide range of anatomies guiding the cannula along safe paths using only bi-planar X-ray input.
\textit{In silico} the policy achieves a 68.5\% first-pass success rate, while the performance decreased to 49.2\% for a separate dataset containing fractured anatomy. Preliminary rollout for planning on real X-ray achieves 40.9\% successful trajectories, demonstrating partial transfer in some cases and offering insight into the method's current capabilities and failure modes. For this preliminary test on real X-ray the model only required two X-ray scans to propose the full trajectory, augmenting each view with simulated tool movement. These results highlight that key elements of intra-operative
decision making can be learned from X-ray input alone. We establish a foundation and benchmark for future efforts towards CT-free guidance systems that could integrate seamlessly into existing workflows.

\section{Methods}\label{sec3}

\begin{figure}[!t]
\centering
\includegraphics[width=1.0\textwidth]{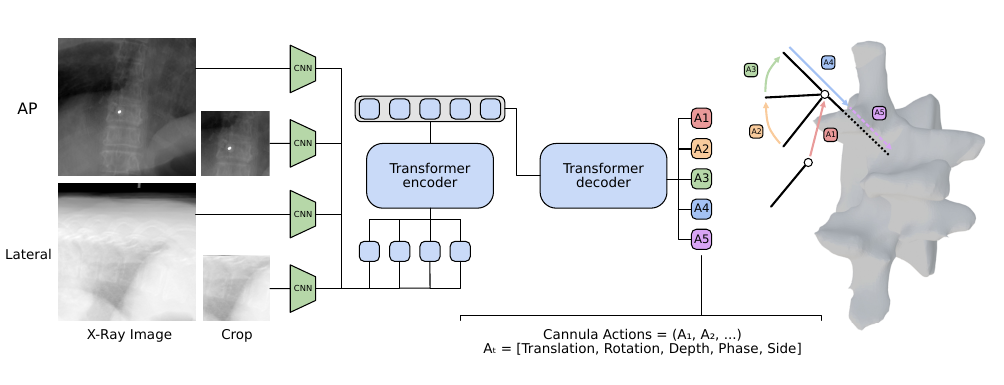}
\caption{\textbf{Model Overview.} Left to right: Inputs consisting of current anterior-posterior (AP) and lateral (LAT) observations are processed via a conditional variational autoencoder. Fine-grain pose adjustments for cannula are predicted as actions to generate final insertion trajectory while modeling surgeon like adjustments.}\label{model_overview}
\end{figure}

To enable surgeon-like spinal trajectory planning directly from bi-planar X-rays we train a transformer-based IL policy using an incremental action representation~\cite{kim_surgical_2024} predicting cannula adjustments from anterior-posterior (AP) and lateral (LAT) radiographs. Fig.~\ref{model_overview} shows a high-level overview of the model during rollout.

\subsection{Data and Preprocessing}

\begin{figure}[!t]
\centering
\includegraphics[width=1.0\textwidth]{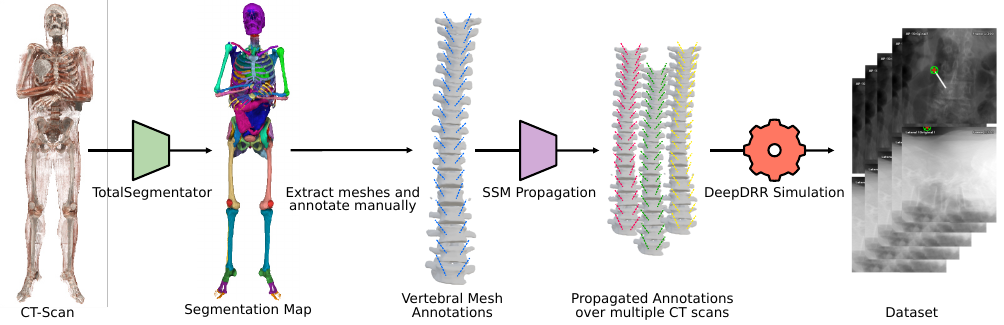}
\caption{\textbf{Data Generation Pipeline.} Left to right: CT-scans from the NMDID~\cite{edgar2020_NMDID} dataset are preprocessed using TotalSegmentator~\cite{wasserthal_totalsegmentator_2023}. Then representative Statistical Shape Models (SSMs) are extracted, manually annotated and propagated over multiple CT scans. Lastly annotations are simulated via DeepDRR to generate our training data.}\label{data_generation}
\end{figure}

The large-scale annotated training data required for training is generated by constructing a realistic simulation environment with automatically derived safe trajectories. The data generation pipeline consists of three main components: (1)~preprocessing of clinical CT data and segmentation; (2)~generation and filtering of safe trajectories; (3)~simulation of bi-planar radiographs. The full pipeline is illustrated in Fig.~\ref{data_generation}.

The foundation for our simulation is provided by a wide range of CT scans from the NMDID dataset~\cite{edgar2020_NMDID}. Each CT is segmented with TotalSegmentator~\cite{wasserthal_totalsegmentator_2023} to obtain 3D vertebrae segmentations. To improve image quality for small fields of view, we increase the CT's voxel resolution by resampling, interpolating, and smoothing tissue volumes~\cite{shaker_synthesizing_2024}. We then assign custom DeepDRR material models~\cite{unberath_deepdrr_2018} with mass densities and material decompositions to the individual anatomical parts.
This produces high-quality radiographs for visualizing small structures such as pedicles with the limited voxel resolution of the CT inputs. In addition, vertebral meshes are generated for efficient trajectory generation and analysis.

\subsection{Trajectory Annotation and Filtering}

Reference trajectories were first manually defined in various vertebral SSMs for representative vertebrae (T1–L5) with left and right pedicle entries. These annotations were propagated across subjects via SSM registration and surface distance-based path interpolation \cite{vijayan_automatic_2019}. To ensure anatomical safety, we automatically filter trajectories with mean cortical wall distance lesser than 1 mm as well as any with cortical breaches based on the existing meshes. For each entry, multiple parallel trajectories were generated, and the trajectory with the greatest mean wall distance after entering was selected.

\subsection{Simulation Environment}

\begin{figure}[!t]
\centering
\includegraphics[width=1.0\textwidth]{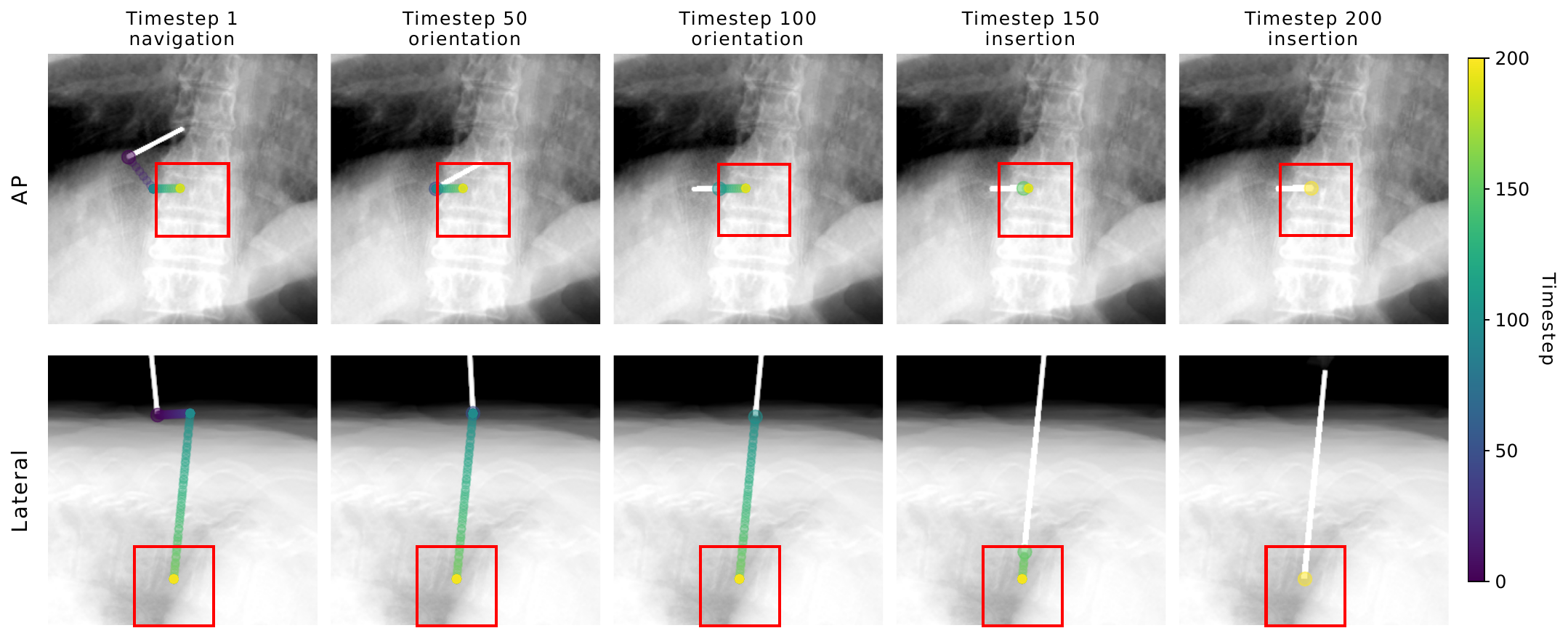}
\caption{\textbf{Example training episode from the NMDID dataset.} Visualization of safe insertion into vertebra T12 via the left pedicle at multiple timesteps. Top: anterior-posterior (AP) view; bottom: lateral (LAT) view, with post-processing applied. Red rectangles mark cropped regions around the target vertebra. Coloured dots indicate the remaining trajectory as projected cannula tip positions.}\label{example_data}
\end{figure}

We implemented the training and testing environment with DeepDRR~\cite{unberath_deepdrr_2018, killen2025fluorosam}. For each trajectory, bi-plane AP and LAT projections are simulated at each navigation timestep to guide the model. The AP views are defined as the average axis between left and right pedicles, while the LAT corresponds to the patient's right-facing direction. Each view is additionally limited to a small anatomical region containing only a few visible vertebrae. To mimic intra-operative variability we applied random view offsets ranging in between $\pm5^{\circ}$ and $\pm2.5\text{cm}$ in the XYZ directions of the CT. The tool is modelled as a cannula used in vertebroplasty procedures which has a length of 14 cm. It is initialized with a random pose near the target vertebra. Each episode consisted of navigation, orientation and insertion phases, yielding 7390 episodes with a length of 200 timesteps across 252 patients. Fig.~\ref{example_data} shows an example of a simulated safe trajectory. It should be noted that even after preprocessing our CT dataset to improve resolution, pedicle corridors remain difficult to see in the simulated X-rays because of the limited input voxel resolution.

\subsection{Policy Representation and Architecture}

We follow the ACT model, a conditional variational autoencoder (CVAE) using a backbone to extract image features and predicting a series of tokens as movement actions, presented in detail by Zhao et al.~\cite{zhao_learning_2023}, with modifications to action representation from SRT~\cite{kim_surgical_2024}. The image backbones were modified to accept four single channel radiographic images per timestep: AP and LAT projections, along with a cropped image region each, around the targeted vertebra. Image features are extracted using a ResNet18 backbone retrained on radiographs. During rollout, image features are fused into a combined representation and processed via the CVAE's transformer decoder. As shown in Fig.~\ref{model_overview} this predicts a series of tokens (action chunks) representing the trajectory, following the training and rollout setup of the original ACT~\cite{zhao_learning_2023}.

Actions are 11-dimensional vectors predicting the delta movement between each timestep. Each action consists of 6-Degrees of Freedom (DoF) for translation and orientation and an additional DoF reserved for the insertion movement as soon as the desired pose is reached and locked in. Additionally we add several flags for redundancy and reproducibility, three phase flags representing navigation, orientation and insertion in a one-hot encoded format, and one binary flag for left or right pedicle entry.

\subsection{Implementation Details}\label{sec_implementation}

The policy is trained on a NVIDIA RTX A6000 for 2000 epochs end-to-end via behavioral cloning without domain randomization, minimizing the L1 error between predicted and reference actions~\cite{zhao_learning_2023}, hyperparameters remain the same from SRT with a modified beta value of 5 \cite{kim_surgical_2024}. We separate the generated episodes randomly as a 70/20/10 split for training, validation and testing. Evaluation is set up with the DeepDRR simulation environment also used for data generation, using the ACT parameters for temporal aggregation of action chunks with a query frequency of 1 to ensure a smooth rollout~\cite{zhao_learning_2023}. 
Generating a single chunk of 20 action steps requires around 150 ms, as for each prediction a new set of bi-planar radiographs has to be simulated. Simulating the full procedure of translation, orientation and insertion then takes around 30 seconds, resulting in approximately 200 aggregated timesteps.

\section{Results}\label{sec4}
We evaluate the proposed policy across multiple categories to characterize both its performance and limitations. First, we assess safety and first-pass success on held-out synthetic cases from the NMDID dataset~\cite{edgar2020_NMDID} using a clinically motivated grading scheme. Second, we quantify geometric accuracy by comparing predicted trajectories against safe reference plans, reporting entry-point and angular deviations. Third, we evaluate generalization to anatomically challenging cases using unseen CTs with vertebral fractures~\cite{loffler_vertebral_2020}. We then analyze sensitivity to initialization and input configuration through ablation studies. Finally, we examine preliminary sim-to-real feasibility via rollouts on a small set of real bi-planar X-rays.

\subsection{Safety and Success Rates on Held-out Synthetic Cases}

\begin{figure}[!t]
\centering
\includegraphics[width=0.9\textwidth]{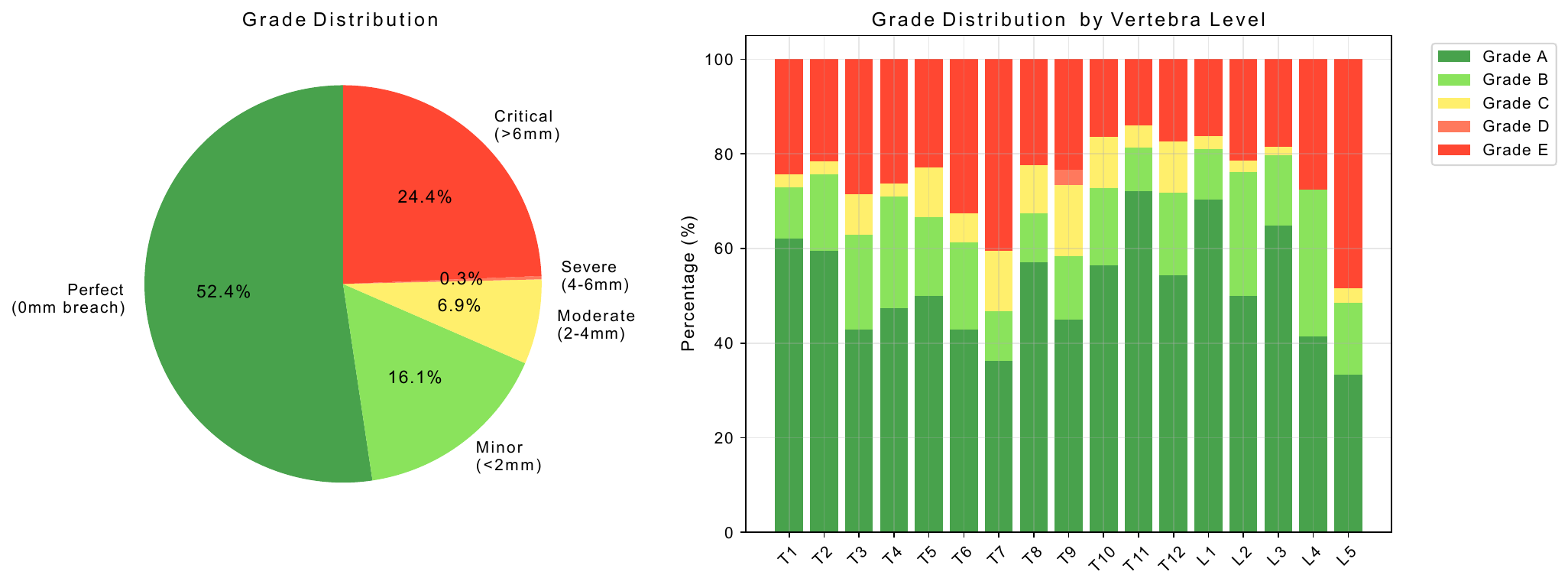}
\caption{\textbf{Safety grading on unseen episodes from NMDID dataset.} Left: Grade distribution of breaches (A: no breach, B: $\le$2 mm, C: 2–4 mm, D: 4–6 mm, E: $\ge$6 mm or extra-pedicular). Right: episodes per grade for each vertebral level. Acceptance is defined as Grades A+B, totaling 68.5\%.}
\label{grade_distribution}
\end{figure}

To evaluate the predicted cannula trajectories, we use a modified version of the Gertzbein–Robbins grading scheme originally developed for pedicle screws~\cite{gertzbein1990accuracy}. Each trajectory is automatically assessed for cortical breaches using segmented vertebral meshes~\cite{wasserthal_totalsegmentator_2023}, with the cannula modelled as a 2~mm diameter cylinder during evaluation. Breaches are categorized by their maximum distance from the cortical surface, with Grades~A and~B ($\leq 2$~mm) considered acceptable.

Testing on 739 previously unseen episodes (10\% of the dataset), predicted trajectories achieved an overall acceptance rate of 68.5\% (Grades A+B) on their first attempt, comprising 52.4\% Grade A (no breach) and 16.1\% Grade B ($\le 2$ mm breach). Failure modes comprised of Grade C (2–4 mm): 6.9\%, Grade D (4-6 mm): 0.3\% and Grade E ($\ge6$ mm or outside of pedicle): 24.4\%, see Fig.~\ref{grade_distribution}. Performance across vertebral levels was generally consistent, with the lower thoracic and lumbar regions showing the highest success rates. Slight drops were observed at T7 and L5, while for the latter these were likely influenced by the NMDID~\cite{edgar2020_NMDID} acquisition setup, where subjects were scanned in a prone position, introducing pronounced lumbar curvature and altered pedicle alignment. Overall, the policy maintained pedicle containment on the first try in the majority of cases, with most breaches limited to minor cortical contact. As pedicle corridors are narrow the grading scheme is sensitive to small entry-point errors, with cortical contact often being caused by minor sub-millimeter entry shifts.

\subsection{Geometric Accuracy Relative to Safe Reference Trajectories}

\begin{figure}[!t]
\centering
\includegraphics[width=0.9\textwidth]{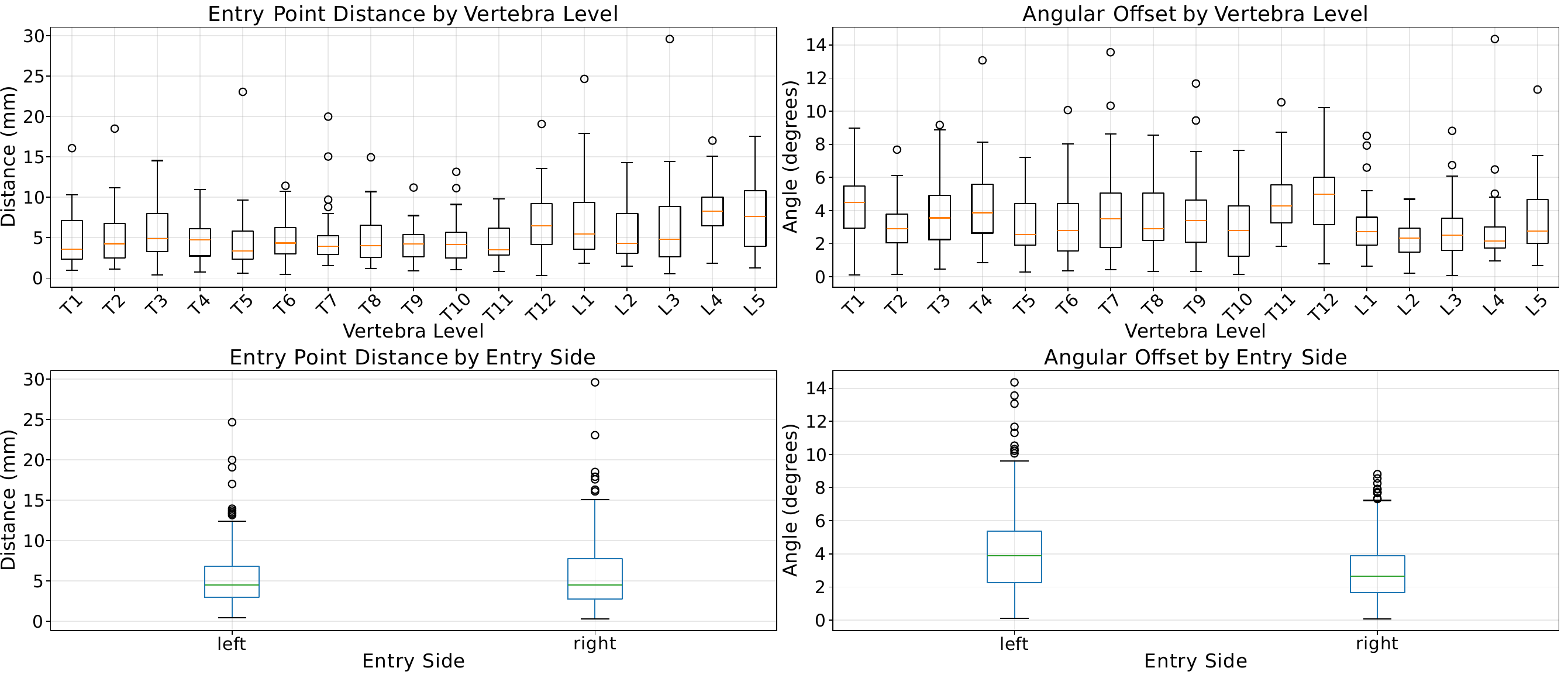}
\caption{\textbf{Geometric agreement with safe generated plans.} Top: entry-point distance (mm, left) and angular offset (degrees, right) grouped by vertebral level. Bottom: the same metrics grouped by pedicle side. Boxes show the inter-quartile range (IQR) with the median line; whiskers denote 1.5$\times$IQR; points are outliers.
}
\label{entry_angle_offsets}
\end{figure}

We compare the trajectories predicted by our method and SSM-derived generated safe plans reporting the geometric error as shown in Fig.~\ref{entry_angle_offsets}. Unsuccessful trajectories not entering the pedicle or vertebra are excluded. Across vertebral levels, mean entry-point error is  $5.46 \pm 3.67 \text{mm}$, while mean angular offset is $3.53 \pm 2.19 ^{\circ}$. When comparing by pedicle side, right-sided entries have a slightly larger entry-point error spread than left-sided entries. Extreme outliers for entry points are episodes with a failed insertion.

\begin{figure}[!t]
\centering
\includegraphics[width=1.0\textwidth]{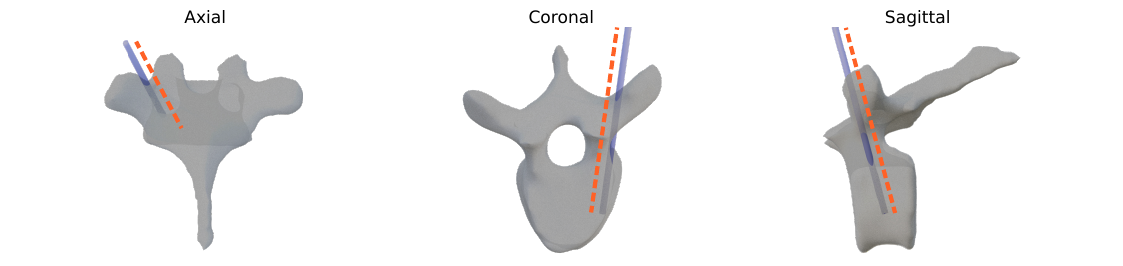}
\caption{\textbf{Example of failed insertion (Grade E).} Left to right: Axial, coronal, and sagittal views of a predicted trajectory (blue) compared to the expert reference (orange) for pedicle access into T6. Angular offset is $3.31 ^{\circ}$ and entry point distance is $9.21 \text{mm}$ compared to expert trajectory.}
\label{visual_inspection_offset}
\end{figure}

Previous work introducing SSMs for automatic trajectory generation reported an average entry-point deviation of approximately $2.5 \text{mm}$ and average angular difference of $3.5^{\circ}$~\cite{vijayan_automatic_2019}. While our policy's angular offset lies within these expected ranges, the stronger deviation in entry-point placement indicates that this is the primary limiting factor to overall performance, as visualized in Fig.~\ref{visual_inspection_offset}.

Overall, median angular offsets across both pedicle sides remain low, suggesting that orientation estimates are generally reliable. In contrast, entry-point placement contributes disproportionately to residual error and Grade~E failures. Because the policy operates iteratively, small deviations during early navigation outside of the body, can compound and significantly reduce the feasible corridor for later corrective actions, making recovery in further orientation and insertion steps difficult. 

\subsection{Generalization to Fractured Anatomy}

\begin{figure}[!t]
\centering
\includegraphics[width=0.9\textwidth]{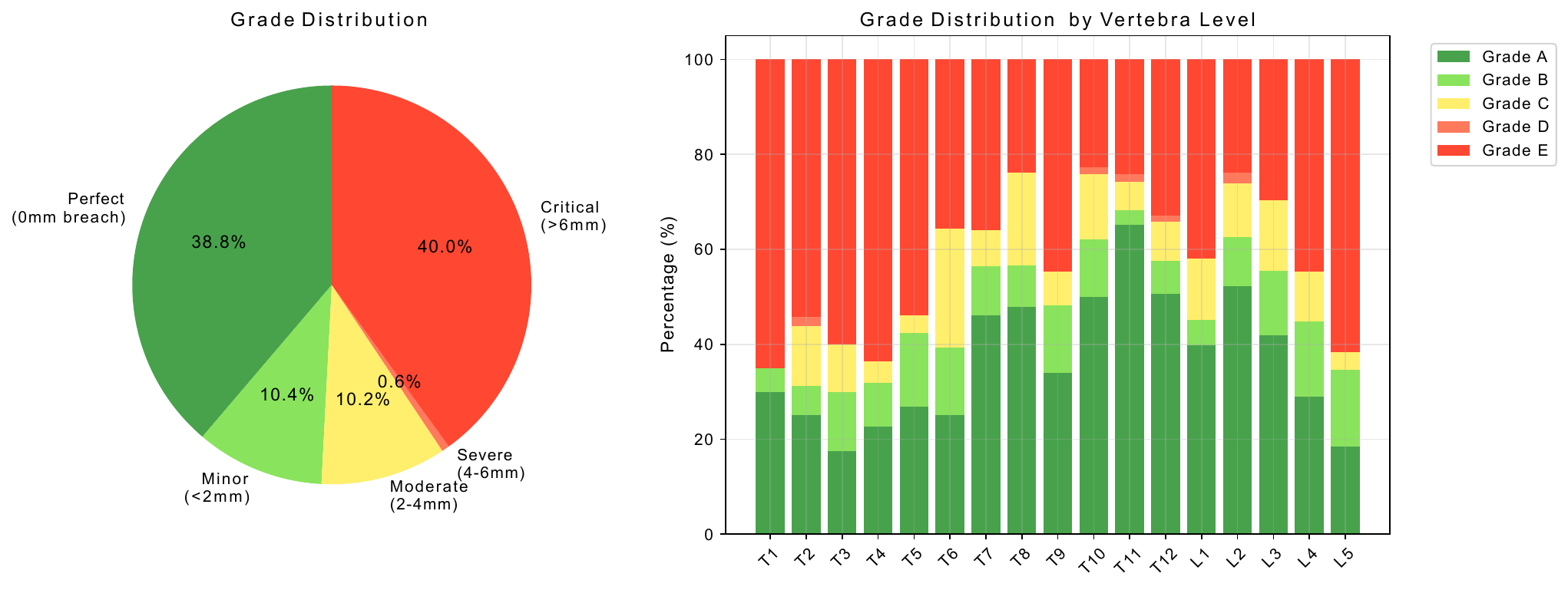}
\caption{\textbf{Safety grading on unseen cases from Verse19 dataset.} Left: Grade distribution of breaches (A: no breach, B: $\le$2 mm, C: 2–4 mm, D: 4–6 mm, E: $\ge$6 mm or extra-pedicular). Right: episodes per grade for each vertebral level. Acceptance is defined as Grades A+B, totaling 49.2\%.}
\label{grade_distribution_verse19}
\end{figure}

\begin{figure}[!t]
\centering
\includegraphics[width=1.0\textwidth]{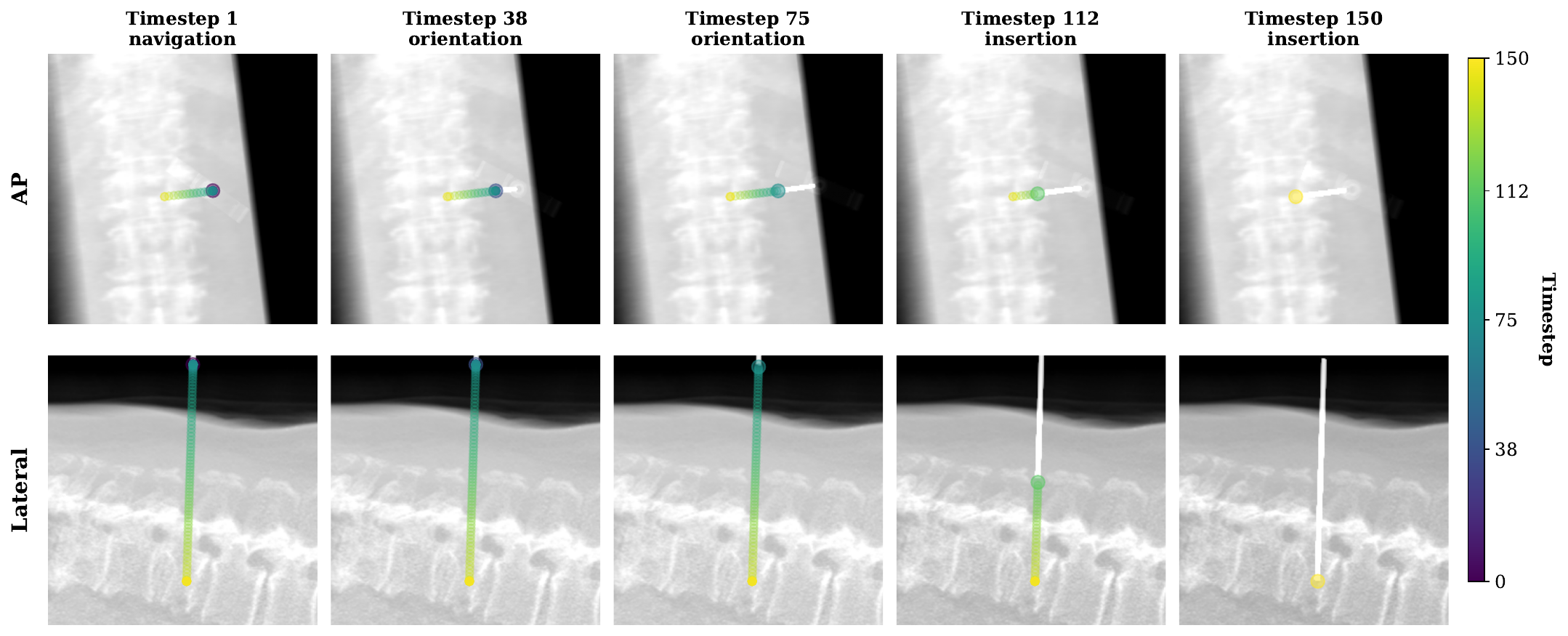}
\caption{\textbf{Predicted episode with fractured vertebra (Verse19).} Visualization of predicted insertion into fractured vertebra L1 with Grade A via the right pedicle at multiple timesteps. Top: anterior-posterior (AP) view; bottom: lateral (LAT) view, with post-processing applied. Coloured dots indicate the remaining trajectory as projected cannula tip positions.}\label{example_fractured}
\end{figure}

To further evaluate generalization, we tested the policy on a set of 63 previously unseen CTs containing vertebral fractures \cite{loffler_vertebral_2020}, yielding 962 additional episodes. Unlike the training dataset (NMDID \cite{edgar2020_NMDID}), this test set exhibits an imbalanced vertebral distribution with notably fewer samples in the upper thoracic region (T1–T9). On this dataset, the policy achieved an overall acceptance rate of 49.2\% (Grades A+B), reflecting a 16.2\% drop compared to performance on held-out synthetic cases. 

As detailed in Fig.~\ref{grade_distribution_verse19}, performance degradation was most pronounced in the highest thoracic levels (T1–T5), likely due to smaller vertebra and anatomical variation, limiting the feasible corridor for the cannula trajectory. This effect increases when introducing variations like fractures, as shown in Fig.~\ref{example_fractured}, further reducing the valid space. The unbalanced distribution of vertebral levels, with reduced data availability at higher thoracic levels, also weakens the evaluation of these levels. Lastly a major limiting factor impacting generalization is the low resolution of CT volumes used for data generation, limiting the effective simulated X-ray resolution, which is a major factor in the successful rollout on higher vertebral regions and anatomically challenging cases.

\subsection{Ablations and Policy Behavior}

\begin{table}[!t]
\centering
\caption{
Ablation results across input configurations. Metrics reported as mean $\pm$ SD. Acceptable trajectories are Grade A+B. \textit{Baseline configuration} and \textit{Center Initialization} use AP, LAT and Cropped views. \textit{Baseline} has a random starting pose, while \textit{Center Init.} starts from a fixed pose above the center of the vertebra. \textit{AP-only}, \textit{LAT-only} and \textit{Without Cropped} all have a random starting pose. LAT-only was unable to execute any insertions into the vertebra.}
\label{tab:ablations}
\begin{tabular}{lccc}
\toprule
\textbf{Config} & \textbf{Angular Offset ($^{\circ}$)} & \textbf{Entry-Point Dist. (mm)} & \textbf{Acceptable (\%)} \\
\midrule
Baseline & $3.53 \pm 2.19$ & $5.46 \pm 3.67$ & 68.5 \\
Center Init. & $6.32 \pm 4.21$ & $6.76 \pm 4.70$ & 41.0 \\
\midrule
AP-Only & $5.52 \pm 4.31$ & $8.84 \pm 5.16$ & 17.6 \\
LAT-Only & $-$ & $-$ & 0.0 \\
Without Cropped & $5.09 \pm 3.78$ & $7.63 \pm 4.71$ & 42.2 \\
\bottomrule
\end{tabular}
\end{table}

To assess sensitivity of the model to initialization pose we evaluate the policy on initializing from a central pose above the vertebra instead of a randomized starting pose. We also perform ablations of the model using only AP and a cropped region, only LAT and a cropped region, and AP and LAT without any cropped guidance, to assess the importance of the bi-planar views with cropped regions for guidance.

As shown in Table \ref{tab:ablations}, removing one view degraded performance significantly. The AP-only policy exhibited increased angular and entry-point error and frequently breached the bottom cortical wall. LAT-only input was unable to execute insertions into the vertebra and could not be measured in comparison to expert trajectories. Without cropped regions for guidance the model failed to find the relevant small pedicle insertion corridors consistently, while initializing from a central position above the vertebra lead to an increased likelihood of central foramen breaches. All of the reported ablations also achieve higher success rates toward the lumbar vertebral levels.

\subsection{Evaluation on Real X-rays}

\begin{figure}[!t]
\centering
\includegraphics[width=1.0\textwidth]{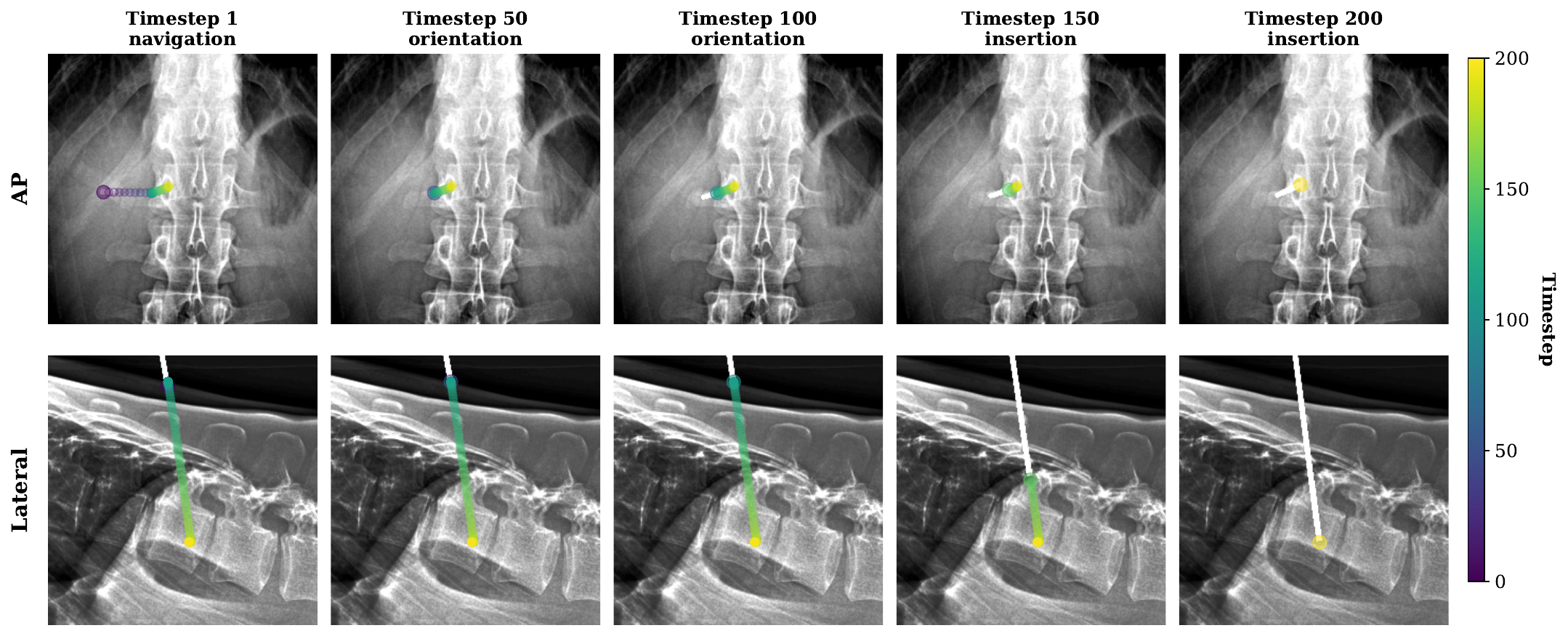}
\caption{\textbf{Predicted trajectory on real X-Ray augmented with simulation of cannula.} Visualization of predicted insertion through the left pedicle at multiple timesteps. Image was generated by blending simulation of cannula movement with real bi-plane X-Ray views. Top: anterior-posterior (AP) view; bottom: lateral (LAT) view, with post-processing applied. Coloured dots indicate the remaining trajectory as projected cannula tip positions.}\label{example_real_trajectory}
\end{figure}

We evaluated the policy on two bi-planar (AP/LAT) X-ray pairs from the BUU-LSPINE dataset~\cite{klinwichit2023buu} to assess sim-to-real performance. After manual alignment and cropping around four lumbar vertebrae, left and right pedicles were targeted with five rollouts each, yielding 80 simulated insertions. Planning requires only one AP and one LAT acquisition, which are augmented with simulated cannula motion. The cannula is rendered using DeepDRR by initializing it from a random pose and overlaying it onto the real radiographs, an example can be seen in Fig.~\ref{example_real_trajectory}.

Outcomes were then graded manually from bi-planar views due to the absence of CT ground truth. Fourteen rollouts failed to initiate insertion and were excluded. Manual grading of the remaining 66 attempts was performed in two rounds: an initial conservative assessment by two researchers with domain knowledge of vertebroplasty, followed by an independent review by a clinical specialist not involved in model development. A summary of grading outcomes is shown in Table~\ref{tab:real_xray_grading}, achieving a 40.9\% success rate during the specialist review. The higher acceptance rate assigned by the clinical specialist likely reflects a clinically informed tolerance to minor trajectory deviations, whereas the researchers applied a more conservative assessment due to limited clinical context. 

We observed three dominant failure modes on this dataset: (1) breach of the pedicle corridor, resulting from entry-point deviation; (2) failure to initiate insertion, associated with poor initial alignment; and (3) attempted post-insertion pose updates. Post-insertion updates were seen in 15.2\% of cases and often caused initially acceptable trajectories to fail, which resulted in unrealistic over-advancement as well as loss of containment.

We also note that the right pedicle was less visible in these cases, likely due to patient pose and projection geometry, degrading performance. Imperfect alignment, introduced by manual preprocessing, further reduces the viable corridor by shrinking the pedicle corridor intersection of both views. These results suggest preliminary sim-to-real transfer in some cases, while highlighting clear limitations, like entry-point precision and the need to restrict post-insertion updates in any application.

\begin{table}[!t]
\centering
\caption{Manual grading results for real X-ray evaluation. Trajectories were classified as \textit{Clinically Acceptable}, \textit{Borderline} (acceptable but would likely need correction), or \textit{Unacceptable}. Grading was performed independently in two rounds.}
\label{tab:real_xray_grading}
\begin{tabular}{lccc}
\toprule
\textbf{Reviewer} & \textbf{Clinically Acceptable} & \textbf{Borderline} & \textbf{Acceptance (\%)} \\
\midrule
Researchers & 9 & 14 & 34.8 \\
Clinical Specialist & 21 & 6 & 40.9 \\
\bottomrule
\end{tabular}
\end{table}

\section{Discussion}

This study investigates the feasibility of policy-driven planning for X-ray-guided spine procedures under extensive simulation of diverse anatomies and imaging conditions. Across simulated cases, the proposed IL policy achieves a 68.5\% acceptance rate on unseen synthetic cases (n=739 randomized trials) and 49.2\% on an unseen dataset including fractured anatomies (n=962 randomized trials), without relying on CT or registration. 
Evaluation on real clinical X-rays, without retraining, yields 40.9\% acceptable trajectories (N=2 subjects,  n=80 randomized trials), indicating that partial sim-to-real transfer is possible in some cases, while substantial gaps remain.

Performance is primarily limited by entry-point localization under sparse views, compounded by limited DRR fidelity and the iterative rollout that reduces recoverability after insertion onset. Failures on real X-rays indicate a need for explicit phase constraints and broader training distributions. Rule-based constraints, such as locking translational and rotational DoF when the desired pose for insertion is reached, could improve clinical safety for cases in which over-correction is the main cause for breaches in future systems. However such constraints would require an additional layer of supervision during the transition between phases, evaluating the correct moment to engage such a lock.
Overall these results indicate that while open-loop planning from sparse bi-planar X-ray is feasible, it remains sensitive to initialization and phase structure. Although the high update rate used during planning raises concerns about radiation exposure, our real X-ray results demonstrate that full trajectories can in some cases be planned only from two initial views using tool augmentation.

In contrast to previous planning approaches for vertebral procedures, our method learns the full cannula insertion trajectory and movement, rather than predicting only a final target pose with a screw shape. Prior work in this domain assumes access to CT data with preoperative planning, with intra-operative execution guided by registration-based navigation systems~\cite{vijayan_automatic_2019, siemionow_autonomous_2021, huang_current_2021}, which our method does not require. Given the current performance limitations, advantages of ACT are its direct usage of intra-operative data streams and live planning capabilities of the full insertion trajectory, with the possibility of starting from arbitrary states~\cite{zhao_learning_2023}. Additionally, while simulating a full procedure currently requires around 30 seconds, the iterative nature of generating the full prediction horizon through a set of aggregated chunks, can immediately integrate movement cues, like patient respiration, into open-loop planning.

Our results indicate that while transformer-based policies such as ACT perform well with high-frame rate video input from multiple angles, their assumptions break down in sparse view X-ray imaging. In such 2D medical modalities we lack texture, depth and context. DRR simulation further increases this difficulty because of limited resolution. This reflects the general difficulty of reasoning about precise 3D geometry from sparse 2D inputs alone and highlights the need for models capable of integrating global anatomical cues across low contrast images.

\section{Conclusion}

We presented a feasibility study toward CT-free, vision-based planning for vertebroplasty using imitation learning on bi-planar X-rays. Results from extensive simulation and limited real data show that elements of surgeon-like alignment behavior can be learned directly from sparse radiographic input, while also revealing clear limitations in accuracy, robustness, and sim-to-real transfer. These findings highlight both the promise and constraints of applying ACT-style policies to medical imaging domains in which standard visual assumptions do not hold. 

Rather than serving as a standalone autonomous solution, the proposed framework is best viewed as an assistive system that could support intra-operative trajectory planning and alignment using data already acquired during fluoroscopy-guided procedures, with partial human supervision remaining essential for entry-point selection and insertion start. Incorporating real-time human corrections and demonstrations into training provides a direct path to improve robustness, reducing entry-point error, and better constrain post-insertion behavior. With stronger anatomical priors, vision-transformer-based encoders, and domain-adaptive training, future systems may achieve improved spatial understanding and accuracy, moving closer to practical CT-independent guidance.

\backmatter

\noindent\textbf{Supplementary information:}
The online version contains supplementary material in the appendix.

\noindent\textbf{Funding:} This work was funded in part by NIH R01 EB036341 and Johns Hopkins Internal Funds.

\section*{Declarations}
\noindent\textbf{Competing interests:} The authors have no competing interests to declare that are relevant to the content of this article.

\noindent\textbf{Ethical approval:} This article does not contain studies with human participants performed by any of the authors.

\noindent\textbf{Informed consent:} This article does not contain patient data collected by any of the authors.

\bibliography{sn-bibliography}

\newpage

\begin{appendices}

\section{Policy Learning Details}\label{secA1}

We train an imitation policy using Action Chunking with Transformers (ACT)~\cite{zhao_learning_2023}. At each timestep $t$, the policy observes four single-channel radiographs:

\begin{equation}
o_t = \{I_t^{AP}, I_t^{LAT}, I_t^{AP^c}, I_t^{LAT^c}\},
\end{equation}
where $(\cdot)^c$ denotes a crop around the target vertebra. The policy predicts a short \emph{action chunk} of length $H$:
\begin{equation}
\hat{\mathbf{a}}_{t:t+H-1} = \pi_\theta(o_t).
\end{equation}

\paragraph{Action representation}
Each action $a_t \in \mathbb{R}^{11}$ encodes an incremental update in the local frame of the cannula:
\begin{equation}
a_t =
\big[
\Delta \mathbf{p}_t,\ \Delta \boldsymbol{\omega}_t,\ \Delta d_t,\ \mathbf{s}_t,\ b_t
\big],
\end{equation}
with translation $\Delta \mathbf{p}_t\in\mathbb{R}^3$, rotation $\Delta \boldsymbol{\omega}_t\in\mathbb{R}^3$, insertion increment $\Delta d_t\in\mathbb{R}$, phase flag $\mathbf{s}_t\in\{0,1\}^3$ (navigation/orientation/insertion, one-hot encoding), and a binary side flag $b_t\in\{0,1\}$ (left/right pedicle).

\paragraph{Behavioral cloning loss (ACT)}
We train end-to-end by behavioral cloning, minimizing the L1 loss between predicted and demonstrated actions over the chunk:
\begin{equation}
\mathcal{L}_{BC}(\theta)=
\mathbb{E}_{(o_t,\mathbf{a}^\star)\sim\mathcal{D}}
\left[
\left\|
\pi_\theta(o_t) - \mathbf{a}^\star_{t:t+H-1}
\right\|_1
\right].
\end{equation}

Following ACT~\cite{zhao_learning_2023}, the model is implemented as a conditional variational autoencoder (CVAE) to encourage temporally coherent chunk predictions, with a KL term weighted by $\beta$ ($=5$).

\paragraph{Rollout}
At inference, we query the policy every timestep (query frequency $=1$) and apply temporal ensembling to smooth the predicted action sequence. The executed action corresponds to the first aggregated action of the current predicted chunk~\cite{zhao_learning_2023}.

\end{appendices}

\end{document}